\documentclass[twoside,english]{IEEEtran}
\usepackage[T1]{fontenc}
\usepackage[latin9]{inputenc}
\usepackage{graphicx}
\usepackage[justification=centering]{caption}
\usepackage{comment}

\makeatletter

\providecommand{\tabularnewline}{\\}

\usepackage{babel}

\makeatother

\usepackage{babel}
\begin{document}
\title{Deep Clustering based Boundary-Decoder Net for Inter and Intra Layer
Stress Prediction of Heterogeneous Integrated IC Chip}

\author{\IEEEauthorblockA{Kart Leong Lim, Ji Lin\\
 Institute of Microelectronics (IME), \\
 \textit{Agency for Science, Technology and Research (A{*}STAR),}\\
 2 Fusionopolis Way, Innovis \#08-02, Singapore 138634, Republic of
Singapore \\
 \{limkl, ji\_lin\}@ime.a-star.edu.sg}}

\maketitle
\begin{abstract}
High stress occurs when 3D heterogeneous IC packages are subjected
to thermal cycling at extreme temperatures. 
Stress mainly occurs at the interface between different materials. We investigate stress image using latent space representation which is based on using deep generative model (DGM). However, most DGM approaches are unsupervised, meaning they resort to image pairing (input and output) to train DGM. Instead, we rely on a recent boundary-decoder (BD) net, which uses boundary condition and image pairing for stress modeling. The boundary net maps material parameters to the latent space co-shared by its image counterpart. Because such a setup is dimensionally wise ill-posed, we further couple BD net with deep clustering. To access the performance of our proposed method, we simulate an IC chip dataset comprising of 1825 stress images. We compare our new approach using variants of BD net as well as a baseline approach. We show that our approach is able to outperform all the comparison in terms of train and test error reduction.     
\end{abstract}

\section{Introduction}
 Advanced packaging of IC chips \cite{iyer2016heterogeneous, lau2022recent, mahajan2002emerging, lau2023chiplet} involves complex heterogenous integration. The integration between different structures and materials, e.g. die, UF and RDL, inside IC chips, is prone to fracture when subjected to thermal cycling \cite{blish1997temperature, lau1993thermal, pang2001thermal, niu2009study, tsai2004investigation, zhang2006modeling}. Stress prediction \cite{park2007predictive, hsieh2004artificial, suhir2009predictive} assist IC designers to analyse thermal induced crack in their designs by simulating stress contour image at the cross-sectional planes of IC chip at different layers. Simulating the stress contour of the IC chips involves the advanced knowledge of different material properties, and they affect stress prediction differently at inter and intra layer level. The material parameter comprising different components, the design variables and process parameters all come together to form an input vector for a corresponding stress contour image. However, there are several challenges when performing stress prediction on IC chips. First, when preparing a design of
experiment (DOE) for stress prediction, the number of simulation cases extracted using finite element analysis (FEA) \cite{vandevelde2014chip, shen2016finite, jorgensen2020overview} is a problem. In particular, the number of cases for a full factorial dataset is dependent on L levels and P parameters at each layer. From finite element simulation perspective, full factorial DOE at each layer (intra layer) is not feasible. Second, deep generative model can elevate the need for a full factorial DOE using latent space representation. However, label is unknown in the latent space and inverse modeling of the latent space for stress prediction is not well established. Third, unlike intra layer modeling, different cross-section (inter layer) of the IC chip yield very different stress contour plots. This is due to the heterogenous integration \cite{lee2020multi, lu2009materials} of complex interfaces between die, underfield (UF) and redistribution layer (RDL). There is no continuity behavior in the image contour at inter layer representation. While each layer can be represented by a deep generative model, it is not practical to train a different model for each layer. Thus, stress prediction is challenging notably due to inverse modeling of stress prediction and having a different model at each layer. We can depict the complexity of inter and intra layer stress prediction in Fig 2. Recent works related to image based stress prediction have mostly used convolutional neural networks and generative adversarial net. Typically, the neural net takes in the shape, boundary and load conditions as images and outputs a stress image. These works are however not related to IC packages and mainly work on dataset that contains homogeneous 2D structures or material fracture. Also, unlike simple 2D structures, the computational cost of obtaining dataset from finite element simulation is much higher for IC package dataset. Current deep generative model cannot easily extend to both intra and inter layer representation. To cope with heterogeneous latent space from inter layer representation, our strategy is to discretize the latent space into Z layers. Thus, we could address both inter and intra class representation, the former which requires a prior knowledge on selecting the number of layers. We propose using an inverse deep generative model for intra layer stress prediction. The model is based on a recently published boundary-decoder net, which consist of an autoencoder and a boundary net. The autoencoder trains a continuous or homogeneous latent space using image contours. The boundary net is responsible for mapping material parameter as vector inputs into the latent space. After training, inverse modeling on the latent space allows us to reconstruct a stress contour image for an unknown vector input. Using finite element simulation, we ran an IC package dataset of 1825 cases for both inter and intra layer. We train the boundary-decoder net using boundary-decoder loss and reconstruction loss. Whereby each iteration, both supervised and unsupervised training is computed on the latent space. To further improve the performance of boundary-decoder net, we further couple it with deep clustering. As far as we are aware of, this is a novel AI approach which has not been attempted before. Specifically,  for each iteration a clustering algorithm such as Kmeans is re-computed on the latent space, before applying deep clustering loss to update the network. As there are no related work, we develop a baseline method known as the AE+KNN to mimic the inter and intra layer representation of our proposed method. We demonstrate on an inhouse IC chip dataset that our proposed deep clustering based boundary-decoder net significantly outperforms baseline method and boundary-decoder net for stress prediction on heterogeneous integrated IC chips for a two die IC design.  

We survey works related to mechanical stress image prediction 
using deep learning \cite{jiang2021stressgan,yang2021deep,wang2021stressnet,gao2020deep}.
Nie et al \cite{nie2020stress} introduced image pairs $\pi=\left\{ shape\;image,stress\;image\right\} $
of 2D structures with 120,960 samples. The input and output images
are used to train convolutional neural networks (CNN) layer. Instead of CNN, Jiang et al \cite{jiang2021stressgan}
used generative adversarial net (GAN) to perform stress prediction. The
generator provide image pairs to the discriminator to classify whether
it resembles the actual dataset. The discriminator in turn trains the
generator and the process cycle is repeated. Concurrently, Yang et
al \cite{yang2021deep} used conditional GAN (cGAN) to predict stress
image for material composite. cGAN uses a U-net for the generator
and PatchGAN for the discriminator. Following \cite{jiang2021stressgan}, they collected
a dataset of 2000 image pairs $\pi$ for training cGAN. Gao et al
\cite{gao2020deep} predict the stress field of crust and rock formation
using several CNNs in parallel. Each CNN input is an image, and the
output is a X, Y and Z direction stress cube. More recently, Buehler
et al \cite{BUEHLER2022100038} combine CycleGAN with transformer
architecture to train a stress field prediction model when given a
image pairing of image microstructure and stress field. Differently
from \cite{BUEHLER2022100038}, Wang et al \cite{wang2021stressnet} propose future
stress field prediction using temporal independent CNN (TI-CNN) and
bidirectional long short-term memory (Bi-LSTM) net. The TI-CNN takes
microstructure image input but outputs the current feature vector
representing stress at one local region. While Bi-LSTM takes past
and present feature vector to predict future feature vector. Stress
prediction on IC packages is a niche field as it requires cross-discipline
domain knowledge to perform an IC package simulation. While many works
deploy state-of-the-art deep learning for mechanical stress modeling,
their dataset mainly focus on simple 2D structures, unrelated to IC
packages. More importantly, they rely on image pairs $\pi$, which
is not suitable for our case. Our problem requires a multimodal pairing
$\phi=\left\{ parameters\;vector,stress\;image\right\} $ to work.
Currently, all the above survey approaches cannot perform stress prediction
using $\phi$ pairing as their deep learning methods cannot solve multimodal cases.
Also, stress in heterogeneous IC package is very different from homogeneous
2D structure, crust and rock formation or material fracture pattern.

\section{AI dataset}

In this paper, a two dies IC package representative geometry was adopted
to demonstrate the methodology of using AI tool to facilitate AI prediction of stresses at different material interfaces inside
the package. As shown in Fig. 1, two identical rectangle dies were
embedded in the package. Different X-Y, Z-X views of the package structure
are presented in Fig.1. Stress (MPa) occurs at the
bond between different materials when subjected to thermal cycling
($-55$deg to $150$deg). The X and Y axis refers to the plane while
Z and X axis refers to the side view. Silicon (Si) dies are encapsulated by epoxy molding compound (EMC) with overmold on top of dies, i.e. dies are
not exposed. Under the Si dies there is the underfill material layer.
An RDL layer is at the bottom face of the package. 

We focus on the interface between 
\begin{quote}
1) EMC and Die - Overmold layer, \\
 2) Die and UF - UF layer, \\
 3) UF and RDL - RDL layer. 
\end{quote}
which we call the Overmold, RF and RDL layers, as shown in Fig. 2. Train and test cases are generated based on a full factorial DOE with four variables, i.e. EMC modulus, EMC coefficient
of thermal expansion (CTE), die size and gap size between the two
dies. Table I lists the DOE of these four variables. 625 cases ($L=5$,
$P=4$) were generated each layer for $Z=3$ layers using ANSYS
Mechanical for 1825 cases (1500 train and 325 test). The stress result
at each mesh point within the FEA model are exported together with
the associated coordinates for each case. These results are subsequently
used as the training and test data for AI stress predictor model development.
\begin{figure}
\begin{centering}
\includegraphics{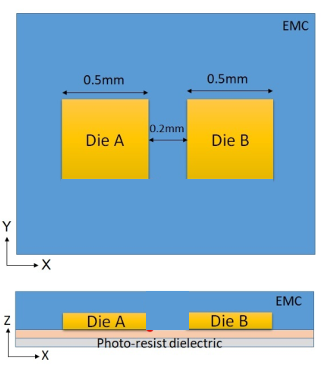} 
\par\end{centering}
\caption{Two dies IC package geometric representation}
\end{figure}

\begin{figure}
\begin{centering}
\includegraphics[scale=0.45]{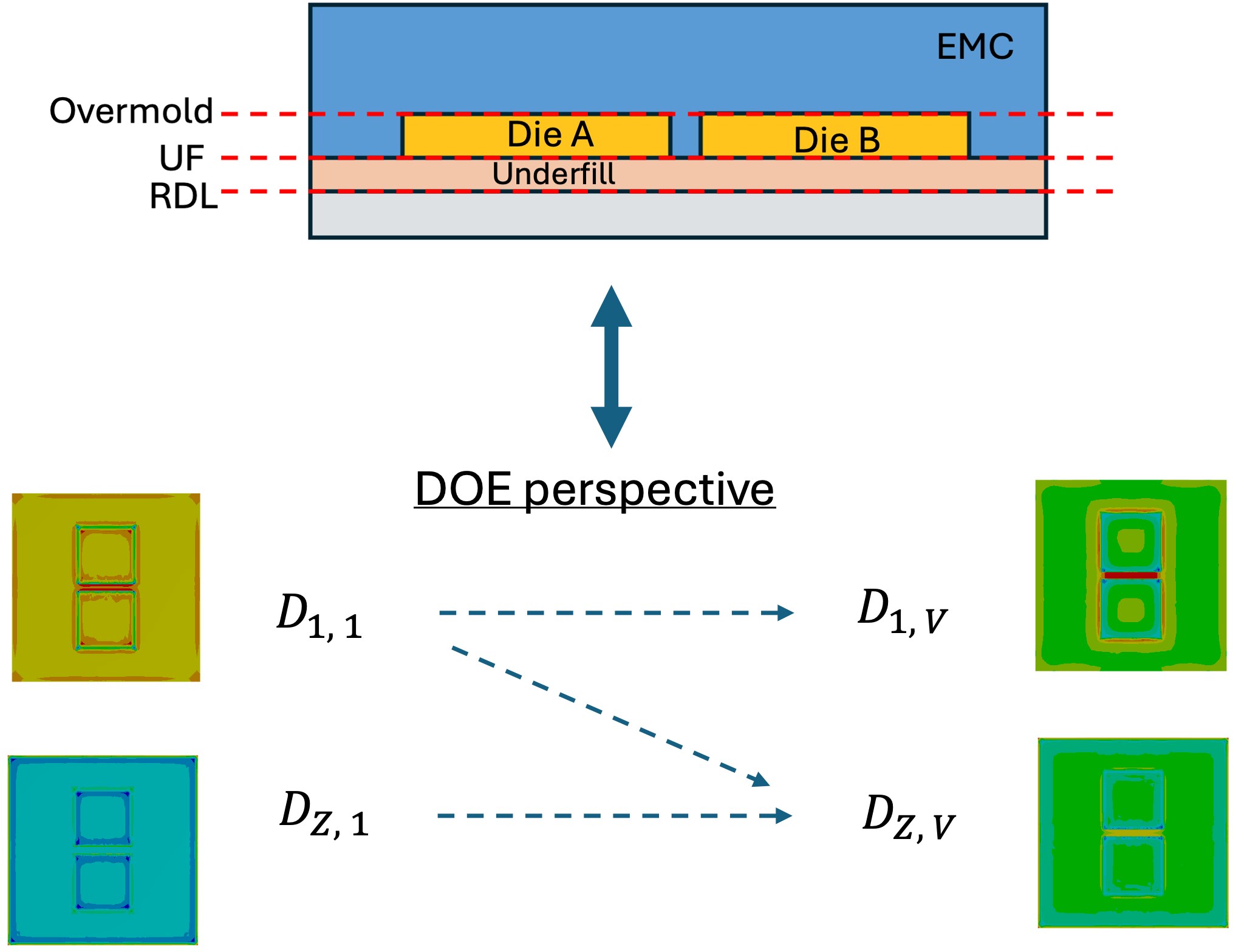} 
\par\end{centering}
\caption{3D representation of stress prediction, using $D_{1...Z,\;1...V}$ where
$Z=3$ layers and $V=L^{P}$ cases.}
\end{figure}

\begin{table}
\caption{DOE for each case using factorial design.}
\begin{centering}
\begin{tabular}{|c||c|c|}
\hline 
Parameters (P)  &  & Levels (L)\tabularnewline
\hline 
\hline 
EMC Modulus {[}GPa{]}  & 5\textasciitilde 30  & 5, 11, 17, 23, 30\tabularnewline
\hline 
EMC CTE {[}ppm/C{]}  & 5\textasciitilde 20  & 5, 9, 12, 16, 20\tabularnewline
\hline 
Die Size {[}mm{]}  & 0.5\textasciitilde 1.8  & 0.5, 0.8, 1.2, 1.5, 1.8\tabularnewline
\hline 
Gap between dies {[}mm{]}  & 0.2\textasciitilde 1  & 0.2, 0.4, 0.6, 0.8, 1.0\tabularnewline
\hline 
\end{tabular}
\par\end{centering}

\end{table}

We refer to the example ground truth stress images in Fig 3. using
cases corresponding to different parameters (column) and
different layer (row). In the overmold layer, there are more stress changes
in the images as compared to the UF and RDL layer. We also
show the minimum and maximum plot of each layer in Fig 4. These values
suggest that the intensity distribution is more consistent for RDL
and UF across their respective 625 cases but changes more abruptly for overmold. This is because the parameters in Table 1 (e.g. die size)  mainly affects stress at the overmold layer. Parameters such as UF or RDL thickness which may affect stress at UF or RDL layer is not present in our study. 

\begin{figure}
\begin{centering}
\includegraphics[scale=0.25]{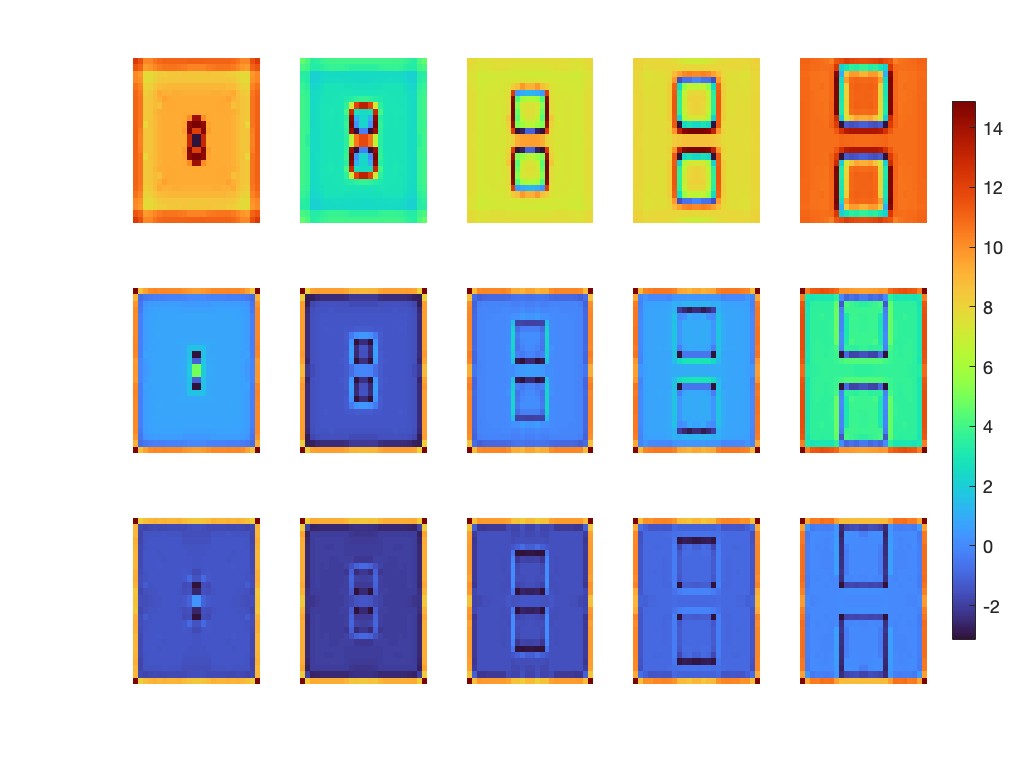} 
\par\end{centering}
\caption{Stress images corresponding to different parameters (column) and at
different layers (row).}
\end{figure}

\begin{figure}
\begin{centering}
\includegraphics[scale=0.5]{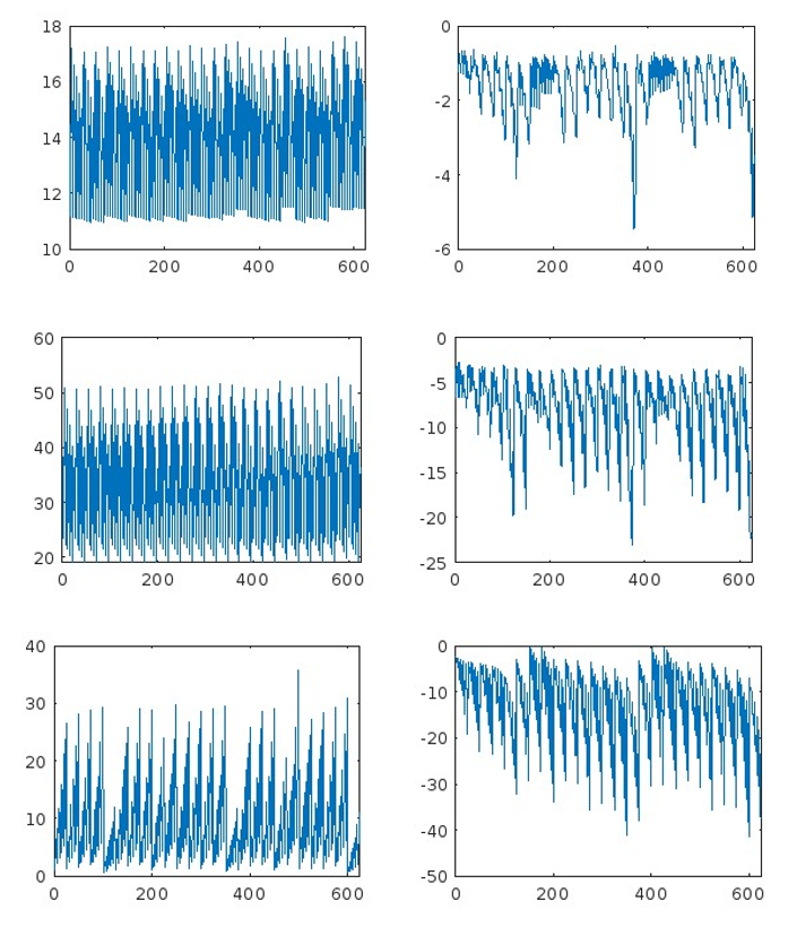} 
\par\end{centering}
\caption{Min and max range for RDL (top row), UF (middle), Overmold (bottom)}
\end{figure}

\section{AI Methodology}

\begin{figure*}
\begin{centering}
\includegraphics[scale=0.5]{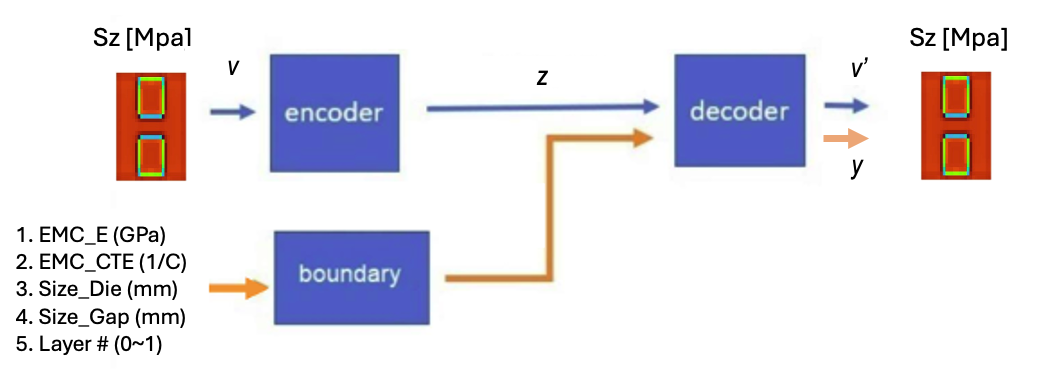} 
\par\end{centering}
\caption{Architecture of Boundary Decoder net}
\end{figure*}

\subsection{Proposed approach using BD}

A recent method known as Boundary Decoder (BD) net \cite{lim2024boundary}
as shown in Fig. 5. (yellow path) first maps low dimensional parameters
representing boundary conditions, to the latent space $z$ of an autoencoder,
and lastly the decoder generates an image. Essentially, BD net behaves
like an inverse prediction model. The problem statement of inverse
prediction \cite{lim2024boundarydecodernetworkinverseprediction} is quoted: 
\begin{quote}
"Given that a regression model can model $1..m$ samples of electrostatic
field as $V_{1...m}$ and corresponding boundary condition parameter
as output $BC(V_{1..m})$. Inverse prediction then tries to recover
input $V'$ when presented with an output $BC(.)$." 
\end{quote}
Inspired by \cite{lim2024boundarydecodernetworkinverseprediction}, we treat material parameters
as boundary conditions\footnote{We use the naming convention in boundary-decoder net \cite{lim2024boundarydecodernetworkinverseprediction}.
Strictly speaking, material parameters are not related to
boundary conditions found in physics models. Instead, the parameters
refer to the physical geometry of the IC chip. } for generating stress image from latent space. This approach requires
offline learning by training the BD net architecture using BD loss,
$L_{3}$ in eqn (1), where $y$ and $T$ are the network image output
and the groundtruth image respectively. The input to the boundary
net are the material parameters, $vec_{train}=\left[EMC_{E},EMC_{CTE},Size_{Die},Size_{Gap},Layer\right]$.
Using $vec_{train}$, groundtruth images and $L_{3}$, we update only
the weights of the boundary and decoder net.
 
\subsection{Proposed approach using AE+BD}

The abovementioned BD approach may be ill-posed due to mapping input
dimensions as low as four parameters to a high dimensional output
such as a 26x26 image. In the original BD net \cite{lim2024boundarydecodernetworkinverseprediction},
it is shown that for capacitor modeling, by jointly training BD with
autoencoder (AE), it significantly reduces the error of the generated
BD image output. Specifically, we train the full network of Fig 1.
using both $L_{1}$ and $L_{3}$ loss in eqn (1). This has the effect
of training the decoder using different objectives, both using reconstruction
loss and using BD loss. Since the test case mainly
uses BD for output, we regularize the weight update of $L_{1}$ using
$\lambda_{1}$ for the decoder. In eqn (1), $\left(V,T\right)$ and $\left(V',y\right)$
refers to the groundtruth and decoder output. We used different terminology
to distinguish between BD and AE for the loss variables.

\subsection{Proposed approach using DC+BD}

While the $L_{1}$ loss of AE improves the performance of BD, we seek
to further constraint the latent space by using deep clustering \cite{lim2025deepclusteringusingadversarial}.
The goal of deep clustering \cite{lim2025deepclusteringusingadversarial} is to regularize
the encoder using statistical method. It actively recomputes $Kmeans$
and penalize latent samples for being far apart from their nearest $K$
cluster centers. Specifically, deep clustering constraint all the training
images according to their similarities in the latent space. This similarity
is determined by 1) the number of clusters predetermined, 2) feature extraction by the encoder and 3) the intensity distribution of the images. The advantage of this approach is that we
can train the encoder can better capture the intra layer representation in the latent space. We train the deep clustering based boundary-decoder (DC+BD) net by using
all the loss functions found in eqn (1). In eqn (1), $\eta^{*}$ refers
to the nearest $K$ cluster center and $z$ refers to the encoder output. When training
DC+BD, the encoder is trained using $L_{2}$ with hyperparameter $\lambda_{2}$ and by $L_{1}$ with  $\lambda_{1}$. The decoder is trained by $L_{1}$ and $L_{3}$. While the boundary net is trained by $L_{3}$.

\begin{equation}
\begin{array}{c}
Loss=\lambda_{1}L_{1}+\lambda_{1}L_{2}+L_{3}\\
=\left\{ -\frac{\lambda_{1}}{2}\left(V-V'\right)^{2}\right\} +\left\{ -\frac{\lambda_{2}}{2}\left(\eta^{*}-z\right)^{2}\right\} +\left\{ -\frac{1}{2}\left(T-y\right)^{2}\right\} 
\end{array}
\end{equation}

\subsection{Baseline approach using AE+KNN}

We also introduce a baseline for stress image prediction, using AE
and K-Nearest-Neighbor (KNN). The objective is found in eqn (2). Using
training images, we train an AE to represent images in the latent
space. This is stored in the form of an array $latent_{train}$, corresponding
to their input vector counterparts $vec_{train}$. For an unseen test
input vector, $vec_{test}$ is compared against $vec_{train}$ using
KNN to select an optimal training sample $vec_{train}^{*}$. Then,
the optimal $vec_{train}^{*}$ samples which corresponds to its $latent_{train}^{*}$
sample counterpart, is in turn fed to the decoder to reconstruct an
image, $V'$. The autoencoder is trained using $L_{1}$ loss in eqn
(1) to update the weights of the encoder and decoder. The cons of
this approach is that both AE and KNN training are disjointed and
the method requires a large training dataset to make up for the the
lack of offline learning involved. The pro is that this method is fast to compute, apart from requiring a large
memory storage for $latent_{train}$.

\begin{equation}
\begin{array}{c}
vec_{train}^{*}=KNN\left(vec_{test},vec_{train}\right)\\
\\latent_{train}^{*}=F_{vec\rightarrow latent}\left\{ vec_{train}^{*}\right\} \\
\\V'=F_{latent\rightarrow decoder}\left\{ latent_{train}^{*}\right\} 
\end{array}
\end{equation}

\section{Experiments}

\subsection{Training and testing error comparison}

\begin{figure}
\begin{centering}
\includegraphics[scale=0.5]{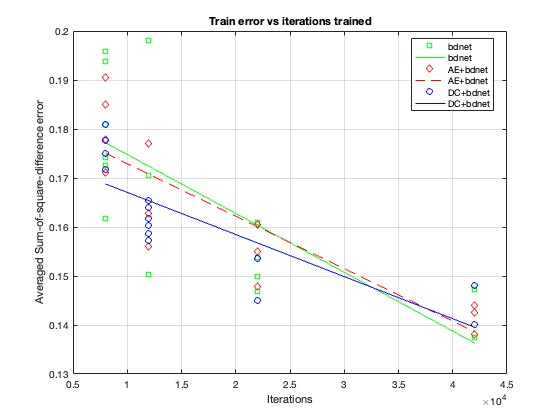}
\par\end{centering}
\caption{Scatter points and regression lines for 1500 train samples.}

\begin{centering}
\includegraphics[scale=0.5]{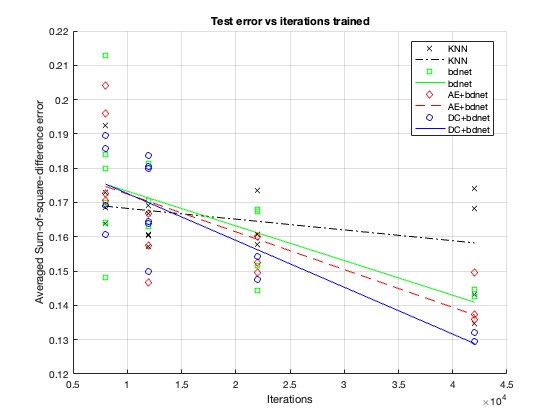}
\par\end{centering}
\caption{Scatter points and regression lines for 375 test samples.}
\end{figure}

\begin{table}[t]
\caption{Comparison of Boundary-Decoder (BD) variants vs baseline}
\begin{centering}
\begin{tabular}{|c|c|c||c|}
\hline 
 & train error  & test error  & train time\tabularnewline
\hline 
\hline 
AE+KNN(baseline)  & -  & 0.1589$\pm$0.0186  & 30\textasciitilde 40s\tabularnewline
\hline 
BD (proposed)  & 0.1424$\pm$0.0069  & 0.1437$\pm$0.0014  & 8\textasciitilde 25s\tabularnewline
\hline 
AE+BD (proposed)  & 0.1416$\pm$0.0031  & 0.1409$\pm$0.0074  & 25\textasciitilde 50s\tabularnewline
\hline 
DC+BD (proposed)  & 0.1441$\pm$0.0056  & 0.1309$\pm$0.0019  & 400\textasciitilde 600s\tabularnewline
\hline 
\end{tabular}
\par\end{centering}

\end{table}

We make a comparison of our proposed Boundary-Decoder variants (BD,
AE+BD, DC+BD) with baseline (AE+KNN) in Table II and Fig. 6. Each
scatter point refers to the sum-of-square-difference error between
groundtruth and generated image for a particular iteration termination
run. We rerun each method for at least 3-5 times from initial random
weights with different iteration termination runs (8000, 12000, 22000,
42000) to show their scatter plots, along with their optimal linear
regression lines (using ``fitlm'' in Matlab). In the train error
graph in Fig. 6, we observe that BD and AE+BD have slightly on-par
regression lines across, while DC+BD leads early on but converge similarly
to the other two methods. AE+KNN is not valid for this comparison.
We support this finding in Table II, where all variants of BD have
similar average train error after 42000 iterations.

When we look at the test error in Fig 7, it is evident that baseline
AE+KNN appears to have limited error reduction, while all variant
of BD outperforms AE+KNN after training for at least 12000 iterations.
Notably, DC+BD outperforms all the methods by a large margin at 42000
iterations. Table II numerically shows AE+BD and DC+BD outperform
BD by 1.9\% and 8.9\%. Moreover, DC+BD leads AE+KNN by 17.6\%. This
suggest that while the train error of the supervised trained BD is
on par with AE+BD and DC+BD, BD generalize poorer than DC+BD when
tested on unseen samples. We attribute this performance gain by the
unsupervised training of AE+BD and DC+BD. In particular, DC+BD achieved
this by using a more compact latent space known as deep clustering,
to minimize intra-cluster variation.

In terms of computational speed in Table II, BD is the fastest, followed
by AE+BD, AE+KNN and lastly DC+BD. The expensive cost in DC+BD is
mainly due to the load of recomputing $Kmeans$ on all the training
samples in the latent space each training iteration.

\subsection{Visual comparison of proposed methods}

Fig 8. to Fig. 10. are using larger die sizes at the upper end of
0.5\textasciitilde 1.8 in Table I. We can visualize the generated
train (top row) and test (bottom row) images of BD, AE+BD and DC+BD
at different layers. For overmold, there is distinct contrast between
the foreground and background. This is because the material parameters
in Table I are directly affecting the geometry in the overmold layer.
The parameter effect is less significant for the UF layer and RDL
layer as they are further away from the two dies. The physical location
of the dies is illustrated in Fig 2. In terms of the metric error
using sum-of-square-difference (ssd), we observed that DC-BD mostly
outperforms the two other variants of BD for all the layers. AE+BD
appears to be poorer than BD. We also note for RDL, it is much harder
for the 3 methods to capture stress as the intensity difference between
foreground and background is not significant.

For the small die size comparison in Fig 11. to Fig. 13, we observe
all variants of BD appear to have image artifact in the generated
images for the overmold layer. We suspect this is could be due to
limited training samples. Overall for small dies comparison, we observe
that BD visually outperforms both AE+BD and DC+BD.

\begin{figure}
\begin{centering}
\includegraphics[scale=0.45]{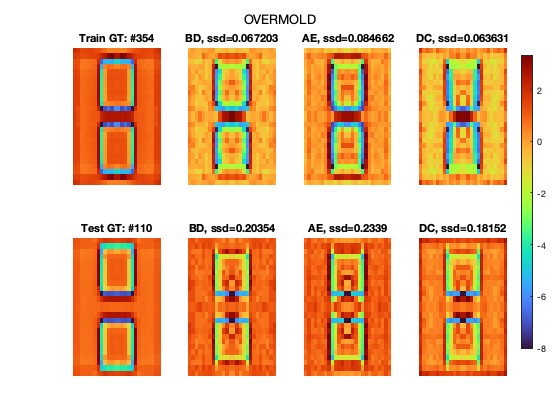} 
\par\end{centering}
\begin{centering} 
\caption{Large die size for overmold layer}
\par\end{centering}
\begin{centering}
\includegraphics[scale=0.45]{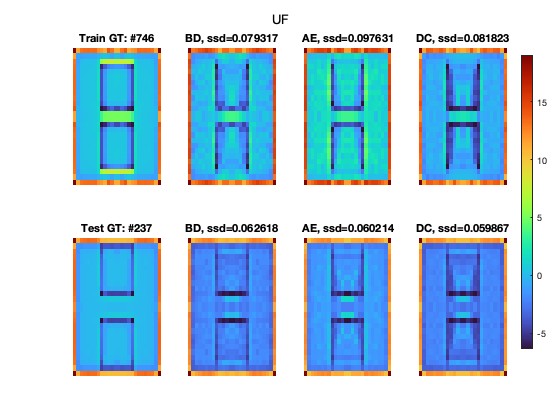} 
\par\end{centering}
\begin{centering}
\caption{Large die size for UF layer}
\par\end{centering}
\begin{centering}
\includegraphics[scale=0.45]{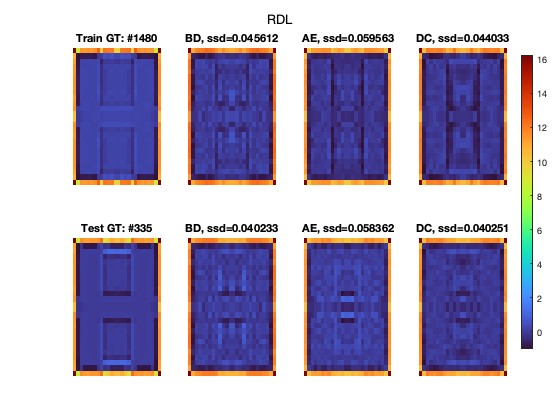} 
\par\end{centering}
\centering{}\caption{Large die size for RDL layer}
\end{figure}

\begin{figure}
\begin{centering}
\includegraphics[scale=0.45]{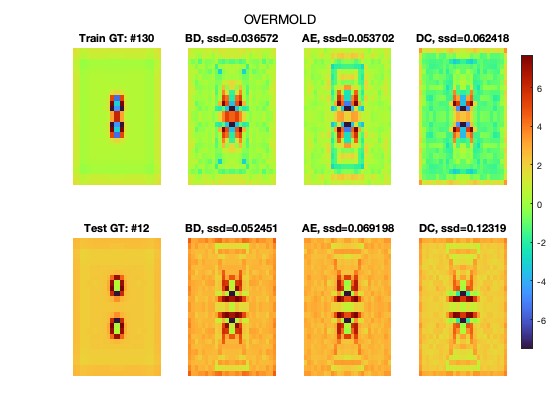} 
\par\end{centering}
\begin{centering}
\caption{Small die size for overmold layer}
\par\end{centering}
\begin{centering}
\includegraphics[scale=0.45]{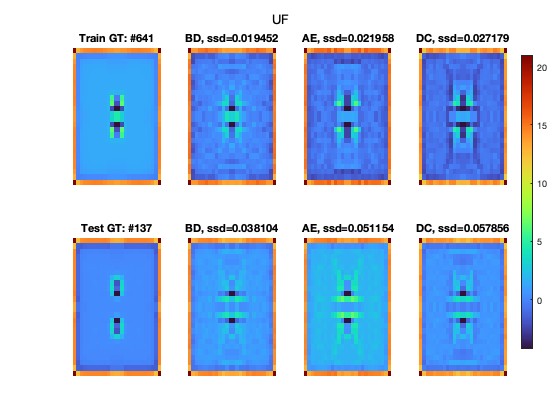} 
\par\end{centering}
\begin{centering}
\caption{Small die size for UF layer}
\par\end{centering}
\begin{centering}
\includegraphics[scale=0.45]{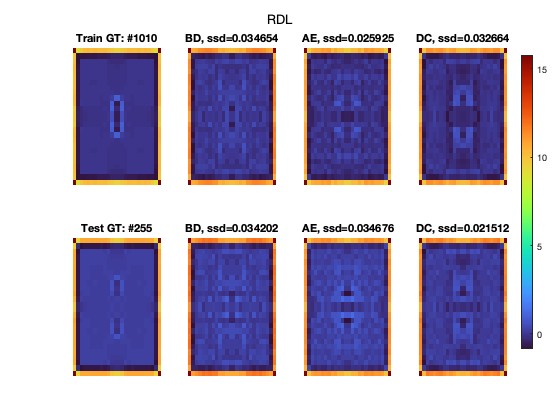} 
\par\end{centering}
\centering{}\caption{Small die size for RDL layer}
\end{figure}

\section{Conclusion}

We introduce stress prediction for a 2 dies IC chip. To cope with
different material parameters and different interface
layers of the IC chip, we represent stress prediction as a latent
space problem. The problem with latent space analysis is that the
samples are unlabeled. To overcome this, we use boundary-decoder net
which jointly trains labels (continuous value) with image in the latent space. The labels
of the boundary-decoder net refer to material parameters. To generate
an image, the latent sample of the label is fed into the decoder.
To further improve the decoder capability, we enhance the training
using deep clustering. We developed a stress prediction dataset of
1825 samples where a comparison is made between baseline and proposed
variants of boundary-decoder. We demonstrate that boundary-decoder
is successful at performing stress prediction on our inhouse dataset.
In future, we would like to explore how stress prediction can be generalized
to varying number of dies.

\section{Acknowledgement} 
This work is supported by "Predictive Package Integrity and Reliability for Multi Chiplet Heterogeneous Integration Products"

 \bibliographystyle{IEEEtran}
\addcontentsline{toc}{section}{\refname}\bibliography{allmyref}

\end{document}